\documentclass[letterpaper, 10 pt, conference]{ieeeconf} 

\usepackage[T1]{fontenc}

\IEEEoverridecommandlockouts                             

\usepackage[protrusion=false]{microtype}
\usepackage{graphicx}
\usepackage{booktabs} 
\usepackage{multirow}
\usepackage{caption} 
\usepackage{siunitx} 
\usepackage[table]{xcolor} 
\usepackage{stfloats} 
\usepackage{capt-of}
\usepackage{amsmath}
\usepackage{colortbl}
\usepackage{amssymb}
\usepackage{tabularx}
\usepackage{textcomp}
\usepackage{pifont}
\usepackage{svg}
\newcolumntype{Y}{>{\centering\arraybackslash}X}
\usepackage{tikz} 
\usepackage{xcolor,colortbl}

\definecolor{rblue}{rgb}{0,0.5,1}
\definecolor{awesome}{rgb}{1.0, 0.13, 0.32}
\definecolor{hollywoodcerise}{rgb}{0.96, 0.0, 0.63}
\definecolor{lasallegreen}{rgb}{0.03, 0.47, 0.19}
\definecolor{hanpurple}{rgb}{0.32, 0.09, 0.98}
\definecolor{green(pigment)}{rgb}{0.0, 0.65, 0.31}

\definecolor{mygray}{gray}{.9}

\definecolor{nbarrier}{RGB}{255, 100, 0}
\definecolor{nbasegreen}{RGB}{0,100,0}
\definecolor{headergray}{gray}{0.95}
\definecolor{rowgray}{gray}{1}
\definecolor{ourgreen}{RGB}{235,250,235}
\newcommand{\cg}[1]{\cellcolor{rowgray}#1}
\newcommand{\cog}[1]{\cellcolor{headergray}#1}
\newcommand{\co}[1]{\cellcolor{headergray}#1}

\usepackage{booktabs}

\makeatletter
\let\NAT@parse\undefined
\makeatother
\usepackage[pagebackref=false,breaklinks=true,colorlinks,bookmarks=false]{hyperref}
\hypersetup{colorlinks=true,linkcolor={red},citecolor={hanpurple},urlcolor={magenta}}

\title{\LARGE \bf
NOVA: Next-step Open-Vocabulary Autoregression for 3D Multi-Object Tracking in Autonomous Driving
}

\author{Kai Luo$^{1,*}$, Xu Wang$^{2,*}$, Rui Fan$^{3}$, and Kailun Yang$^{1,\dag}$
\thanks{This work was supported in part by the National Natural Science Foundation of China (Grant No. 62473139), in part by the Hunan Provincial Research and Development Project (Grant No. 2025QK3019), and in part by the State Key Laboratory of Autonomous Intelligent Unmanned Systems (the opening project number ZZKF2025-2-10).}
\thanks{$^{1}$The author is with the School of Artificial Intelligence and Robotics and the National Engineering Research Center of Robot Visual Perception and Control Technology, Hunan University, Changsha 410082, China (email: kailun.yang@hnu.edu.cn).}
\thanks{$^{2}$The author is with the College of Mechanical and Vehicle Engineering, Hunan University, Changsha 410082, China.}
\thanks{$^{3}$The author is with the State Key Laboratory of Intelligent Autonomous Systems, Tongji University, Shanghai 201804, China.}
\thanks{$^{*}$Equal contribution.}
\thanks{$^{\dag}$Corresponding author: Kailun Yang.}
}

\let\oldtwocolumn\twocolumn
\renewcommand\twocolumn[1][]{%
    \oldtwocolumn[{#1}{
    \begin{center}
    \vskip -4ex
        \centering
        \includegraphics[width=0.86\textwidth]{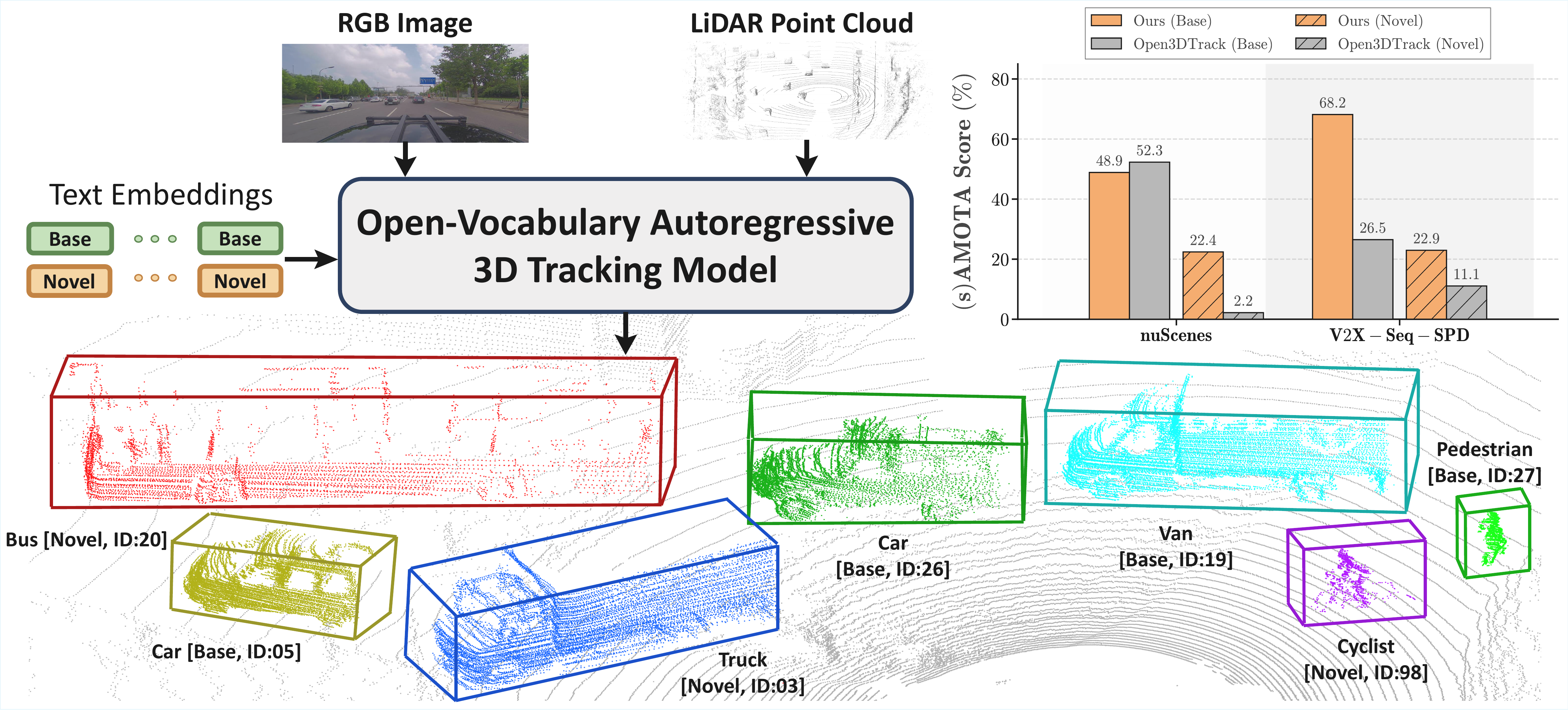}
        \vskip -2ex
        \captionof{figure}{{Overview and performance of our proposed Next-step Open-Vocabulary Autoregression (NOVA) framework for 3D multi-object tracking.
        The framework leverages a Large Language Model (LLM) to autoregressively associate 3D objects across frames using geometric and textual cues. 
        The bottom visualization shows consistent tracking of both Base classes and unseen Novel classes in complex scenes. The top-right chart compares our method with Open3DTrack~\cite{ishaq2025open3dtrack} on nuScenes~\cite{caesar2020nuscenes}, V2X-Seq-SPD~\cite{yu2023v2x}, demonstrating significant improvements in (s)AMOTA, especially for novel categories.}
        }
     \vskip -1ex
    \label{fig:teaser}
    \end{center}
    }]
}

\begin{document}
\maketitle

\thispagestyle{empty}
\pagestyle{empty}


\begin{abstract}
Generalizing across unknown targets is critical for open-world perception, yet existing 3D Multi-Object Tracking (3D MOT) pipelines remain limited by closed-set assumptions and ``semantic-blind'' heuristics. To address this, we propose Next-step Open-Vocabulary Autoregression (NOVA), an autoregressive association formulation that shifts the data association stage from fragmented distance-based matching toward trajectory-conditioned spatio-semantic modeling. NOVA reformulates 3D trajectories as structured spatio-temporal semantic sequences, enabling the simultaneous encoding of physical motion continuity and deep linguistic priors. By leveraging the autoregressive capabilities of Large Language Models (LLMs), we transform the tracking task into a principled process of next-step sequence completion. This mechanism allows the model to explicitly utilize the hierarchical structure of language space to resolve fine-grained semantic ambiguities and maintain identity consistency across complex long-range sequences through high-level commonsense reasoning. Extensive experiments on nuScenes, V2X-Seq-SPD, and KITTI demonstrate the superior performance of NOVA. Notably, on the nuScenes dataset, NOVA achieves an AMOTA of $22.41\%$ for Novel categories, yielding a significant $20.21\%$ absolute improvement over the baseline. These gains are realized through a compact 0.5B autoregressive model. Code will be available at \url{https://github.com/xifen523/NOVA}.
\end{abstract}

\section{Introduction}

\begin{figure}[!t]
    \vspace*{1mm}
    \centering
    \includegraphics[width=0.48\textwidth]{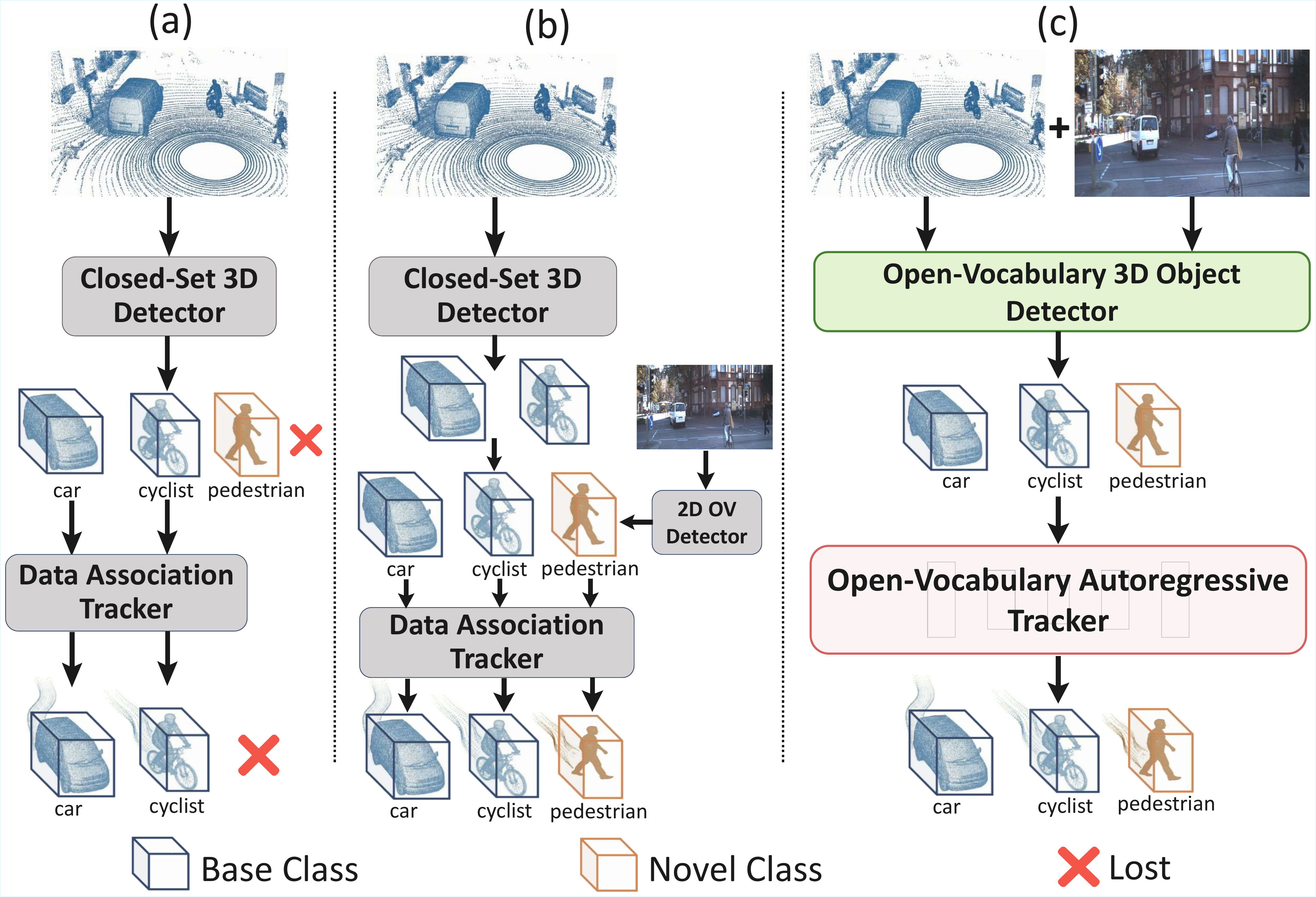}
    \vskip-1ex
    \caption{Comparison of 3D MOT paradigms under open-world category shifts.
    (a) \textbf{Closed-set}: a category-specific 3D detector yields class-limited proposals; unseen objects are suppressed and poorly tracked.
    (b) \textbf{Semi-open-vocabulary}: 2D open-vocabulary predictions are projected onto closed-set 3D proposals; tracking still relies on a downstream association tracker.
    (c) \textbf{Open-vocabulary (ours)}: an open-vocabulary 3D detector outputs labeled 3D detections, and NOVA robustly associates both Base and Novel objects online.}
    \label{fig:paradigm}
    \vskip-3ex
\end{figure}

Autonomous driving operates in complex, safety-critical environments where reliable 3D multi-object tracking (3D MOT) is a fundamental pillar for dynamic scene understanding and downstream planning~\cite{yurtsever2020survey, weng2020ab3dmot}. However, real-world traffic scenarios are inherently open-ended, with various novel objects frequently appearing beyond any predefined training taxonomy~\cite{peri2023towards, bendale2015towards}. 
This mismatch exposes the fundamental limitations of conventional 3D MOT pipelines (as illustrated in Fig.~\ref{fig:paradigm}(a)) built under closed-set assumptions~\cite{caesar2020nuscenes,geiger2012we}: since detectors treat unseen categories as background and suppress them, the system fails to obtain stable proposals to initialize or maintain trajectories, leading to complete tracking failure. These challenges motivate the emergence of Open-Vocabulary 3D Multi-Object Tracking (OV-3D-MOT), which aims to leverage semantic cues to maintain consistent identities for objects beyond fixed category lists~\cite{ishaq2025open3dtrack}.

While 2D open-vocabulary tracking has seen significant progress~\cite{CLIP, li2022grounded, li2023ovtrack, qian2024vovtrack}, its extension to 3D remains hindered by point cloud sparsity, frequent occlusions, and the absence of native 3D open-vocabulary detectors~\cite{weng2020ab3dmot,yin2021center,kim2021eagermot,wang2025mctrack}. Representative efforts, such as Open3DTrack~\cite{ishaq2025open3dtrack} (Fig.~\ref{fig:paradigm}(b)), attempt to bridge this gap via a \emph{post-hoc} strategy, projecting 2D semantics onto 3D proposals from closed-set detectors. However, such a decoupled design leaves geometry generation tethered to closed-set assumptions, resulting in severe localization drift and semantic ambiguity when encountering novel categories. 
Ultimately, this superposition of unreliable geometry and inconsistent semantics bottlenecks the performance of OV-3D-MOT in complex, open-world environments (Fig.~\ref{fig:teaser}).

In light of these limitations, we argue that the core challenge of OV-3D-MOT lies in maintaining long-range logical consistency under uncertainty. 
To this end, we propose NOVA (Next-step Open-Vocabulary Autoregressive 3D Multi-Object Tracker), an association-level reformulation for OV-3D-MOT (Fig.~\ref{fig:paradigm}(c)) that shifts the association score from traditional similarity-based costs to autoregressive spatio-semantic modeling. 
Our key observation is that while traditional trackers treat association as a fragmented task of geometric or visual feature matching, open-vocabulary tracking fundamentally demands a more structured, context-aware reasoning mechanism to navigate an infinite and semantically fluid category space.

NOVA redefines a 3D trajectory as more than just a collection of linked bounding boxes; it is a dynamic ``spatio-semantic sentence''—a sequence that simultaneously follows physical motion laws and linguistic logic. 
By leveraging the autoregressive modeling capabilities of Large Language Models (LLMs), NOVA departs from brittle, semantics-agnostic handcrafted matching rules and instead treats tracking as a principled sequence completion task. This generative approach allows the model to utilize its vast language priors to resolve category ambiguities (\textit{e.g.}, distinguishing between \texttt{SUV} and \texttt{Vehicle}) and maintain identity consistency through high-level semantic common sense, thereby effectively bridging the gap where traditional rules inevitably fail in complex and unconstrained open-world deployments.

At the technical level, NOVA reformulates complex 3D data association as a next-token prediction task over serialized trajectory context. 
This design reduces the reliance on predefined similarity functions by replacing the core association score with an autoregressive decision over serialized inputs.
To build a robust association strategy amidst open-vocabulary uncertainty, NOVA incorporates three core mechanisms: first, a Geometry Encoder maps continuous 3D box states into the word embedding space, supplemented by an IoU auxiliary quality head for regularization to enhance tolerance toward detection noise and localization drift; second, a Hybrid Prompting mechanism is introduced, which masks the specific names of novel categories to force the model to learn intrinsic attribute features rather than memorizing labels, thereby significantly alleviating semantic overfitting to known categories; finally, Hard Negative Mining is employed to specifically sample targets that are spatially proximate but identity-inconsistent, greatly strengthening the fine-grained discriminative power in crowded scenes. 
The deep fusion of these techniques allows NOVA to achieve highly reliable 3D tracking in complex dynamic environments with a minimal parameter cost of only 0.5B.

We conduct extensive evaluations across multiple challenging benchmarks, including nuScenes~\cite{caesar2020nuscenes}, V2X-Seq-SPD~\cite{yu2023v2x}, and KITTI~\cite{geiger2012we}, with results comprehensively validating the superior performance of NOVA. Notably, on nuScenes, NOVA achieves $22.41\%$ AMOTA on novel categories, representing a breakthrough improvement of $20.21\%$ over the Open3DTrack baseline. This significant leap demonstrates that NOVA’s autoregressive paradigm maintains exceptional association robustness by deeply coupling geometric continuity with semantic cues. Furthermore, NOVA consistently outperforms the baseline in sAMOTA on V2X-Seq-SPD and demonstrates remarkable cross-dataset generalization on KITTI.

Our contributions are summarized as follows:
\begin{itemize}
    \item We introduce NOVA, an autoregressive formulation for OV-3D-MOT that casts online 3D data association as next-token prediction over serialized trajectory context, enabling flexible tracking under changing category sets.
    \item We propose a Geometry Encoder with an IoU-based auxiliary quality regression head to align continuous 3D box states with LLM representations, improving geometry-aware association under noisy detections.
    \item We design Hybrid Prompting and hard negative mining to address open-vocabulary semantic uncertainty and strengthen discriminative association, yielding consistent gains especially on novel categories.
    \item Extensive experiments on nuScenes, V2X-Seq-SPD, and KITTI demonstrate state-of-the-art performance of OV-3D-MOT, with consistent gains on novel categories and strong cross-dataset generalization.
\end{itemize}

\section{Related Work}

\subsubsection*{\textbf{Open-Vocabulary Detection}}

Traditional 3D object detection is typically developed under a closed-set taxonomy, limiting generalization to unseen categories~\cite{lang2019pointpillars,shi2020pv,yin2021center}. 
Recent progress in 2D foundation models, including vision-language and grounding detectors~\cite{CLIP,li2022grounded,liu2024grounding,cheng2024yolo}, has enabled recognition beyond fixed label sets and motivates transferring such semantic priors into 3D.
Existing open-vocabulary 3D detection methods navigate two main paradigms: cross-modal representation learning and geometry-based 2D-to-3D propagation. The first approach focuses on aligning point-cloud features with language-supervised visual embeddings or distilling 2D foundation-model knowledge into 3D representations. 
Works such as PointCLIP~\cite{zhang2022pointclip} and OpenScene~\cite{peng2023openscene} exemplify this direction, while newer variants like ImOV3D~\cite{yang2024imov3d} and FM-OV3D~\cite{zhang2024fm} further exploit limited 3D supervision or pseudo-multimodal cues to enhance alignment. 
In contrast, the second paradigm bypasses 3D feature distillation by leveraging 2D open-vocabulary detectors to produce semantic proposals, which are subsequently lifted into 3D via multi-view geometry and clustering. Highlighting the viability of this latter approach, Find n\textquotesingle{} Propagate~\cite{etchegaray2024find} demonstrates how 2D-to-3D lifting can serve as a backbone for urban-scale open-vocabulary~perception.

Despite these advances, LiDAR sparsity and long-tailed distribution shifts~\cite{peri2023towards,bendale2015towards,cao2023coda} can make open-vocabulary 3D detections, especially on novel categories, less accurate and less confident, thereby introducing additional challenges for downstream tracking.

\subsubsection*{\textbf{Open-Vocabulary Tracking}}
Open-vocabulary tracking has been actively studied in 2D, where language-aware representations are integrated into \emph{tracking-by-detection} pipelines (\textit{e.g.}, OVTrack~\cite{li2023ovtrack}) and transformer-based formulations further explore open-vocabulary association under open-set semantics (\textit{e.g.}, OVTR~\cite{li2025ovtr}, VOVTrack~\cite{qian2024vovtrack}). 
In parallel, language-conditioned tracking has been extended to 3D: Wei~\textit{et al.}~\cite{wei2026monocular_3D_tracking} introduce monocular multi-object 3D visual language tracking
with a dedicated benchmark, 
where sentence-level descriptions specify target objects for monocular 3D tracking.
While this line focuses on query-driven 3D tracking conditioned on natural language prompts, OV-3D-MOT targets category-level tracking over open-vocabulary detections and emphasizes robust \emph{data association} under open-set semantic ambiguity.

In contrast, open-vocabulary tracking in 3D remains relatively nascent. 
Open3DTrack~\cite{ishaq2025open3dtrack} is an attempt toward OV-3D-MOT, transferring 2D open-vocabulary semantics onto 3D proposals generated by closed-set detectors~\mbox{\cite{yin2021center,zhu2019class,liu2022bevfusion}}. 
However, its downstream tracking stage largely follows the conventional tracking-by-detection paradigm with predefined association costs and matching solvers. 
Such decoupled designs can be sensitive when open-vocabulary detections become less stable for novel categories, where semantic ambiguity and geometric jitter jointly weaken association cues, motivating alternative association formulations that can more directly incorporate trajectory context and language~priors.

\subsubsection*{\textbf{Autoregressive Modeling for Multi-Object Tracking}}

Autoregressive formulations have long been explored for online MOT to capture temporal dependencies beyond frame-wise matching.
Recurrent autoregressive networks~\cite{fang2018recurrent} couple recurrent dynamics with memory to sequentially update tracking states.
Transformer-based trackers introduce tracking-by-attention with AR-style query propagation, where track representations are updated across frames to maintain identities (\textit{e.g.}, TrackFormer~\cite{meinhardt2022trackformer}). 
For 3D MOT, 3DMOTFormer~\cite{ding20233dmotformer} performs online association on a track--detection bipartite graph via edge classification, and proposes an autoregressive forward strategy to better align training with online inference; TrackFusion~\cite{zhang2025trackfusion} further highlights the value of temporal trajectory modeling for enhancing detection and tracking robustness.
More recently, AR-MOT~\cite{jia2026armot} revisits MOT from a sequence generation perspective within an LLM-style framework, aiming for flexible output formats and task extensibility, while large multimodal models have also been explored for higher-level driving scenario understanding using tracking signals~\cite{ishaq2025tracking_large_models}.

Nevertheless, most prior AR trackers focus on closed-set 2D MOT or geometry-centric 3D MOT, leaving open-vocabulary 3D tracking largely unexplored, where base and novel detections can behave very differently. In this setting, NOVA formulates OV-3D-MOT association as next-token prediction over serialized trajectory context, and improves robustness via geometry-aware embeddings with IoU-quality supervision, hybrid prompting, and hard negative mining.

\section{Methodology}

\begin{figure*}[t] 
    \vspace*{1mm}
    \centering
    \includegraphics[width=0.88\textwidth]{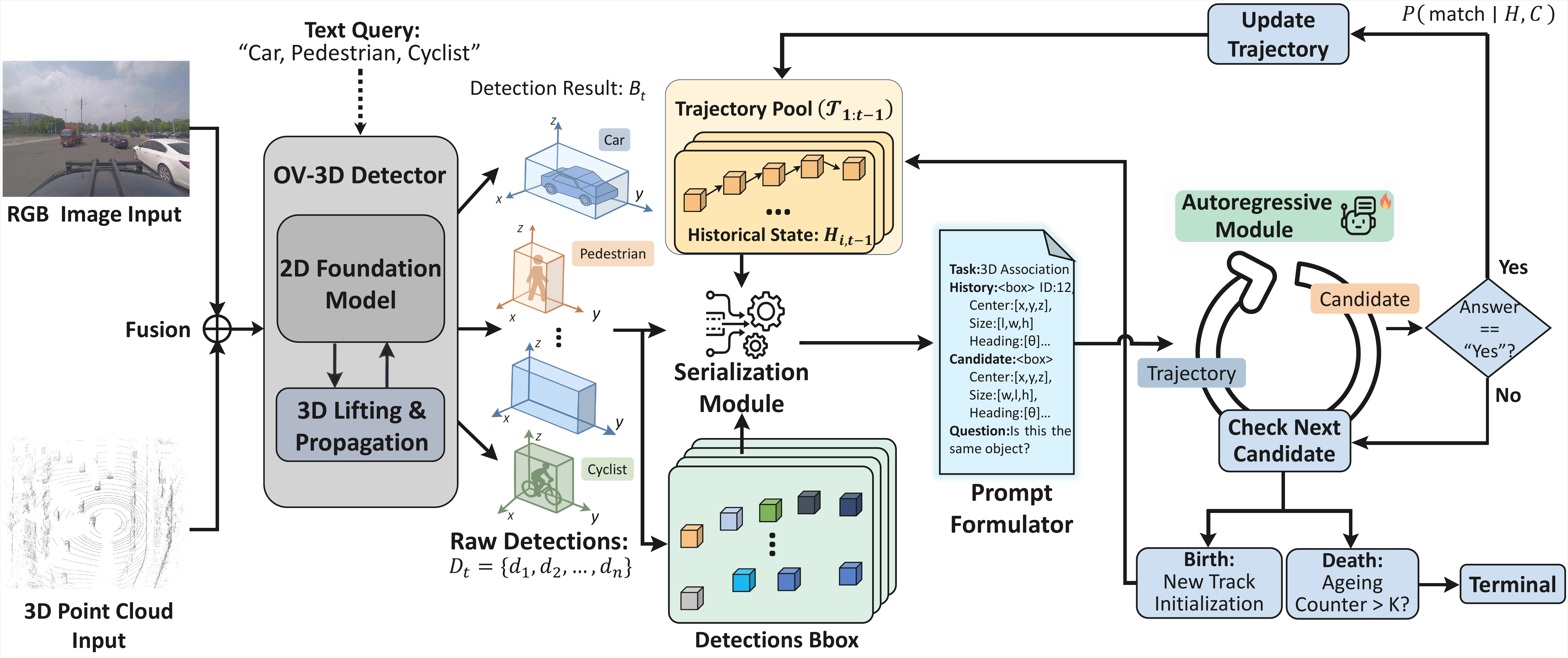} 
    \vskip-1ex
        \caption{\footnotesize
    \textbf{The pipeline of the proposed NOVA framework.}
    (1) \textbf{Open-Vocabulary 3D Detection:}
    Multi-modal inputs are processed to generate detections $D_t$ with open-vocabulary labels.
    (2) \textbf{Serialization \& Hybrid Prompting:}
    A Geometry Encoder projects raw box features $f_{\text{raw}}$ into embeddings $E_{\text{geo}}$. 
    These are interleaved with text using a Hybrid Prompting strategy that explicitly masks novel class labels (\textit{e.g.}, \texttt{Unknown}) to enforce geometric learning.
    (3) \textbf{Autoregressive Association:}
    The LLM predicts a \texttt{Yes} probability for candidate pairs to construct a cost matrix for Hungarian matching, driving the online lifecycle management (Birth/Death) of trajectories.
    }
    
    \label{fig:pipeline}
    \vskip-3ex
\end{figure*}

As illustrated in Fig.~\ref{fig:pipeline}, NOVA follows an online tracking-by-detection framework for open-vocabulary 3D multi-object tracking (OV-3D-MOT). Rather than replacing the entire detection-to-association pipeline, our goal is to reformulate its core \emph{association} stage under open-vocabulary uncertainty.  Specifically, NOVA replaces hand-crafted pairwise similarity costs with an \emph{autoregressive} next-token decision conditioned on serialized trajectory context. To address the inherently unstable geometric cues and inconsistent category semantics in OV-3D-MOT, NOVA introduces three designs: a Geometry Encoder with \texttt{<box>} token injection and IoU-quality supervision, Hybrid Prompting to control semantic visibility, and Hard Negative Mining to strengthen discriminative geometric reasoning.

\subsection{Problem Formulation}
At time step $t$, an open-vocabulary 3D detector outputs a set of detections
$D_t = \{b_t^j\}_{j=1}^{N_t}$.
Each detection is a 3D bounding box
$b_t^j=(x,y,z,l,w,h,\theta,s,c)$,
where $(x,y,z)$ is the box center, $(l,w,h)$ is the size, $\theta$ is yaw, $s$ is the detector confidence, and
$c \in \mathcal{C}_{\text{base}} \cup \mathcal{C}_{\text{novel}}$ is an open-vocabulary label.
The tracker maintains an active trajectory set $\mathcal{T}_{t-1} = \{T_i\}_{i=1}^{M_{t-1}}$, where each trajectory $T_i$ stores a short history of past associated boxes.

Online association aims to decide, for each track--detection pair $(T_i,b_t^j)$, whether they correspond to the same physical object:
\begin{equation}
y_{i,j} \in \{0,1\}, \quad y_{i,j}=1 \Leftrightarrow b_t^j \text{ matches } T_i.
\end{equation}
Under the standard one-to-one constraint, each detection matches at most one active track, and each track matches at most one detection at a time step.
Unmatched detections may initialize new tracks, while unmatched tracks may be kept alive for a short duration to handle occlusion.

In OV-3D-MOT, the association policy must operate under semantic uncertainty: novel categories lack explicit supervision in the upstream 3D detector, leading to noisier detections and frame-to-frame inconsistency in predicted labels and confidences. Consequently, a robust association mechanism should prioritize trajectory-level geometric continuity and incorporate semantics in a controlled, bias-resistant manner, avoiding over-reliance on closed-set category priors.

\subsection{NOVA: Proposed Framework}
NOVA models association as an autoregressive decision.
For each candidate pair $(T_i,b_t^j)$, we serialize the recent history of $T_i$ and append the candidate detection as a query, forming a prompt $\mathcal{P}(T_i,b_t^j)$.
A lightweight LLM is fine-tuned to answer a binary question (match or not) by generating a decision token.
We use the probability of producing the token \texttt{Yes} as the association score:
\begin{equation}
p_{i,j} = \Pr(\texttt{Yes}\mid \mathcal{P}(T_i,b_t^j)).
\end{equation}
This formulation turns association into \emph{context-conditioned} reasoning over a trajectory, rather than a hand-designed combination of independent similarity terms.
Below, we describe three designs that make this autoregressive formulation effective for OV-3D-MOT.

\subsubsection*{\textbf{Geometry Encoder}}
A central challenge is that LLMs operate on discrete token sequences, whereas tracking relies on continuous 3D geometry.
A naive solution is to stringify box states as text (\textit{e.g.}, ``$x=\dots,y=\dots,l=\dots$''), but numerical strings are highly sensitive and can be lossy under tokenization, especially when detections contain small jitters.
NOVA therefore introduces a Geometry Encoder that maps each 3D box into a geometry-aware embedding, which is injected into the LLM input via a dedicated \texttt{<box>} token.

Specifically, for a box $b$ we construct a compact feature vector
\begin{equation}
f_{\text{raw}}=[x,y,z,l,w,h,\text{vol},\theta,s]\in\mathbb{R}^{9},
\end{equation}
where $\text{vol}=lwh$.
The Geometry Encoder projects $f_{\text{raw}}$ to the LLM embedding dimension to obtain $E_{\text{geo}}$.
During prompt construction, each \texttt{<box>} placeholder is replaced by $E_{\text{geo}}$, so that precise geometry is accessible to the LLM as a learned continuous representation rather than brittle numeric text.

\begin{table*}[t]
\vspace*{4mm}
\centering
\caption{\small Comparison of Open3DTrack~\cite{ishaq2025open3dtrack} and NOVA (Ours) on nuScenes~\cite{caesar2020nuscenes}, V2X-Seq-SPD~\cite{yu2023v2x}, and KITTI~\cite{geiger2012we}.}
\label{tab:main_result}

\footnotesize
\renewcommand{\arraystretch}{0.95}
\setlength{\tabcolsep}{2.5pt}

\begin{tabularx}{0.96\linewidth}{c c c c Y Y Y Y Y Y}
    \toprule
    \textbf{Dataset} & 
    \textbf{OV Detector} & 
    \textbf{Method} & 
    \textbf{Cat.} & 
    {sAMOTA} & {AMOTA} & {AMOTP} & {MOTA} & {MOTP} & {MT} \\
    \midrule

    \multirow{4}{*}{\textbf{nuScenes}~\cite{caesar2020nuscenes}} 
    & \multirow{4}{*}{\begin{tabular}{@{}c@{}}Find n\textquotesingle{} Propagate~\cite{etchegaray2024find}\end{tabular}} 
      & \cg{} & \cg{Base} & \cg{--} & \cg{52.30} & \cg{63.70} & \cg{45.00} & \cg{36.70} & \cg{47.27} \\
      & & \cg{\multirow{-2}{*}{Open3DTrack~\cite{ishaq2025open3dtrack}}} & \cg{Novel} & \cg{--} & \cg{2.20} & \cg{126.90} & \cg{2.70} & \cg{62.80} & \cg{30.01} \\
      \cmidrule{3-10} 
      & & \cog{} & \cog{Base} & \co{--} & \co{48.87} & \co{100.09} & \co{43.90} & \co{29.94} & \co{50.51} \\
      & & \cog{\multirow{-2}{*}{NOVA (Ours)}} & \cog{Novel} & \co{--} & \co{22.41} & \co{138.98} & \co{19.60} & \co{55.04} & \co{14.93} \\

    \midrule[\heavyrulewidth] 

    \multirow{8}{*}{\begin{tabular}{@{}c@{}}\textbf{V2X-Seq-SPD}~\cite{yu2023v2x}\\\textbf{(Vehicle Side)}\end{tabular}} 
    & \multirow{4}{*}{\begin{tabular}{@{}c@{}}Find n\textquotesingle{} Propagate~\cite{etchegaray2024find} + \\GroundingDINO~\cite{liu2024grounding}\end{tabular}} 
      & \cg{} & \cg{Base} & \cg{26.50} & \cg{2.50} & \cg{18.34} & \cg{3.01} & \cg{18.40} & \cg{0.60} \\
      & & \cg{\multirow{-2}{*}{Open3DTrack~\cite{ishaq2025open3dtrack}}} & \cg{Novel} & \cg{11.07} & \cg{0.51} & \cg{8.48} & \cg{0.59} & \cg{8.50} & \cg{0.23} \\
      \cmidrule{3-10}
      & & \cog{} & \cog{Base} & \co{68.17} & \co{13.05} & \co{67.19} & \co{35.30} & \co{66.44} & \co{14.37} \\
      & & \cog{\multirow{-2}{*}{NOVA (Ours)}} & \cog{Novel} & \co{22.95} & \co{-141.78} & \co{51.46} & \co{5.30} & \co{51.26} & \co{3.94} \\
     
    \cmidrule{2-10}

    & \multirow{4}{*}{\begin{tabular}{@{}c@{}}Find n\textquotesingle{} Propagate~\cite{etchegaray2024find} +\\YOLO-World~\cite{cheng2024yolo}\end{tabular}} 
      & \cg{} & \cg{Base} & \cg{29.13} & \cg{2.71} & \cg{18.38} & \cg{3.46} & \cg{18.31} & \cg{0.52} \\
      & & \cg{\multirow{-2}{*}{Open3DTrack~\cite{ishaq2025open3dtrack}}} & \cg{Novel} & \cg{15.70} & \cg{0.55} & \cg{8.34} & \cg{0.63} & \cg{8.51} & \cg{0.23} \\
      \cmidrule{3-10}
      & & \cog{} & \cog{Base} & \co{58.88} & \co{8.06} & \co{67.49} & \co{30.80} & \co{67.47} & \co{10.33} \\
      & & \cog{\multirow{-2}{*}{NOVA (Ours)}} & \cog{Novel} & \co{16.01} & \co{-129.85} & \co{50.44} & \co{4.26} & \co{53.21} & \co{2.28} \\

    \midrule[\heavyrulewidth] 

    \multirow{4}{*}{\textbf{KITTI}~\cite{geiger2012we}} 
    & \multirow{2}{*}{\begin{tabular}{@{}c@{}}Find n\textquotesingle{} Propagate~\cite{etchegaray2024find} + \\GroundingDINO~\cite{liu2024grounding}\end{tabular}} 
      & \cog{} & \cog{Base} & \co{93.06} & \co{41.06} & \co{73.38} & \co{80.86} & \co{72.10} & \co{86.99} \\
      & & \cog{\multirow{-2}{*}{NOVA (Ours)}} & \cog{Novel} & \co{12.79} & \co{-104.93} & \co{43.29} & \co{8.73} & \co{40.64} & \co{1.61} \\
     
    \cmidrule{2-10} 

    & \multirow{2}{*}{\begin{tabular}{@{}c@{}}Find n\textquotesingle{} Propagate~\cite{etchegaray2024find} +\\YOLO-World~\cite{cheng2024yolo}\end{tabular}} 
      & \cog{} & \cog{Base} & \co{73.54} & \co{19.58} & \co{73.55} & \co{65.43} & \co{73.25} & \co{69.93} \\
      & & \cog{\multirow{-2}{*}{NOVA (Ours)}} & \cog{Novel} & \co{22.41} & \co{-21.69} & \co{41.18} & \co{11.08} & \co{40.44} & \co{6.45} \\

    \bottomrule
\end{tabularx}
\vskip-4ex
\end{table*}

Beyond representation, OV detectors provide a confidence score $s$, but this score reflects the detector's internal belief and can be poorly calibrated for association: open-vocabulary detections may exhibit compressed or inconsistent confidence distributions, and high-confidence boxes are not always the most association-reliable under occlusion or long-tail categories.
To stabilize learning and encourage $E_{\text{geo}}$ to preserve association-relevant geometry, we add an auxiliary quality head that regresses an IoU-based quality target in $[0,1]$ during training.
This provides explicit supervision on geometric fidelity and offers a principled signal beyond raw detector confidence.

The overall training objective combines decision-token prediction and IoU-quality regression:
\begin{equation}
\mathcal{L}=\mathcal{L}_{\text{gen}}+\lambda_{\text{reg}}\mathcal{L}_{\text{iou}},
\end{equation}
where $\mathcal{L}{\text{gen}}$ is the cross-entropy loss for the binary decision token, and $\mathcal{L}{\text{iou}}$ is an MSE loss between the predicted quality and the IoU target. We set $\lambda_{\text{reg}}=1.0$ in all experiments, treating IoU-quality regression as an equal-weight auxiliary task. Since the IoU target is bounded in $[0,1]$, this setting provides geometric regularization without dominating the main decision objective. Ablations confirm the effectiveness of geometry encoding and IoU-quality supervision.

\subsubsection*{\textbf{Hybrid Prompting}}
A second challenge is the mismatch between training-time semantics and inference-time uncertainty.
If we always condition on category names, the model may learn shortcuts from base categories and generalize poorly to novel categories whose semantics can be ambiguous or unstable.
Hybrid Prompting addresses this by controlling semantic visibility during training:
for base-category samples, we keep explicit class names (\textit{e.g.}, \texttt{Class: Car});
for novel-category samples, we mask the class name with a generic placeholder (\textit{e.g.}, \texttt{Class: Unknown}).
This design explicitly exposes the autoregressive policy to the uncertainty regime expected at inference time, discouraging semantic overfitting and encouraging the model to fall back on geometric and temporal consistency when semantics are unreliable.
Importantly, this does not assume novel detections are reliable; rather, it trains the association policy to remain effective despite semantic ambiguity.

\subsubsection*{\textbf{Hard Negative Mining}}
Finally, association errors are rarely caused by easy background negatives; they are dominated by nearby objects with similar geometry and motion.
Random negative sampling therefore under-trains the decision boundary required for crowded scenes.
NOVA constructs training pairs by sampling positives from matched track--detection pairs, while selecting negatives using hard negative mining:
for each track, we preferentially sample identity-inconsistent detections that are spatially proximate (and thus confusable).
These hard negatives force the model to learn fine-grained geometric discrimination over trajectory context, which directly improves tracking accuracy and trajectory continuity.
We also add mild zero-mean Gaussian noise to the box centers and yaw angles of positive samples to simulate localization noise from upstream detectors.

\subsection{Online Inference and Data Association}
At inference time, NOVA runs strictly online.
Given $\mathcal{T}_{t-1}$ and $D_t$, we first apply lightweight plausibility filtering (\textit{e.g.}, distance-based gating) to discard clearly impossible matches, and then compute $p_{i,j}$ for remaining candidates via the autoregressive model.
We convert these scores into an association cost (\textit{e.g.}, $C_{i,j}=1-p_{i,j}$) and solve a standard bipartite assignment to obtain a one-to-one matching (Hungarian algorithm).
Matched detections update existing trajectories; unmatched detections may spawn new tracks; unmatched tracks are kept active for up to $K$ consecutive missed frames to tolerate short-term occlusion, and are terminated once their age exceeds $K$. In all experiments, we set $K=30$, which provides a practical trade-off between preserving temporarily lost trajectories under occlusion or missed detections and suppressing long-lived false tracks.

Crucially, as illustrated in the pipeline of Fig.~\ref{fig:pipeline} and evidenced by the superior tracking performance in Table~\ref{tab:main_result}, NOVA does not aim to discard the standard one-to-one assignment constraint in online MOT. Instead, it reformulates how the association cost is obtained: the Hungarian algorithm is retained to enforce valid bipartite matching, while each cost is derived from an autoregressive trajectory-conditioned decision rather than category-specific hand-crafted similarity terms. This association-level reformulation integrates geometry and semantics via the three proposed designs, making the framework more adaptive to changing category sets and robust to open-vocabulary uncertainty.

\begin{table*}[t]
    \vspace*{1mm}
    \centering
    \caption{Comprehensive comparison of using different LLM Models on the V2X-Seq-SPD dataset~\cite{yu2023v2x}.}
    \label{tab:backbone_ablation}
    \vskip-1ex
    
    \resizebox{0.98\linewidth}{!}{
        \begin{tabular}{l c ccc cccc c}
            \toprule
            
            \multirow{2}{*}{\textbf{LLM Backbone}} & \textbf{\#Params} & \multicolumn{3}{c}{\textbf{Tracking Performance}} & \multicolumn{4}{c}{\textbf{Stability \& Quality}} & \textbf{Efficiency} \\
            
            \cmidrule(lr){3-5} \cmidrule(lr){6-9} \cmidrule(lr){10-10}
            
             & \textbf{(B)} & \textbf{sAMOTA}$\uparrow$ & \textbf{sAMOTA}$_{novel}\uparrow$ & \textbf{MOTA}$\uparrow$ & \textbf{IDS}$\downarrow$ & \textbf{FP}$\downarrow$ & \textbf{FN}$\downarrow$ & \textbf{MOTP}$\uparrow$ & \textbf{FPS}$\uparrow$ \\
            
            \midrule
            
            \rowcolor{rowgray}
            Phi-3.5-mini-3B & 3.8 & 43.63 & 20.25 & \textbf{21.32} & 266 & 6017 & \textbf{27327} & 59.53 & 0.8 \\
            
            \rowcolor{rowgray}
            Llama-3.2-3B & 3.0 & 45.00 & 22.52 & 21.24 & 131 & \textbf{4355} & 29315 & 59.86 & 1.2 \\
            
            \rowcolor{rowgray}
            Qwen2.5-1.5B & 1.5 & 43.76 & 19.59 & 21.19 & \textbf{115} & 4669 & 28856 & 61.07 & 1.9 \\
            
            \addlinespace[1pt] 
            \rowcolor{headergray}
            \textbf{Qwen2.5-0.5B (Ours)} & 0.5 & \textbf{45.56} & \textbf{22.95} & 20.30 & 182 & 6192 & 27485 & \textbf{58.85} & \textbf{3.4} \\
            
            \bottomrule
        \end{tabular}
    }
    \vskip-3ex
\end{table*}
\section{Experiments}
\begin{table}[t]
    \centering
    \caption{Analysis of the Geometry Encoding Strategy.}
    \label{tab:ablation_geometry}
    \vskip-1ex
    
    \small 
    \begin{tabularx}{0.98\linewidth}{l YY YYY Y Y} 
        \toprule
        \multirow{2}{*}{\textbf{Method}} & 
        \textbf{Geo.} & 
        \textbf{Aux.} & 
        \multicolumn{3}{c}{\textbf{sAMOTA} $\uparrow$} & 
        
        \multirow{2}{*}{\begin{tabular}{@{}c@{}}\textbf{MOTA}\\$\uparrow$\end{tabular}} & 
        
        \multirow{2}{*}{\begin{tabular}{@{}c@{}}\textbf{FN}\\$\downarrow$\end{tabular}} \\

        \cmidrule{4-6}

         & \textbf{Enc.} & \textbf{Loss} & \textbf{All} & \textbf{Base} & \textbf{Novel} & & \\ 
        \midrule
        
        Text      & \ding{55} & \ding{55} & 43.91 & \textbf{68.62} & 19.19 & 19.81 & 28810 \\
        w/o Aux   & \ding{51} & \ding{55} & 45.48 & 68.17 & 22.80 & 19.83 & 29267 \\
        
        \rowcolor{headergray} \textbf{Ours} & \ding{51} & \ding{51} & \textbf{45.56} & 68.17 & \textbf{22.95} & \textbf{20.30} & \textbf{27485} \\
        \bottomrule
    \end{tabularx}
    \vskip-3ex
\end{table}
\subsection{Experiment Setups}
\subsubsection*{\textbf{Datasets and Metrics}}
We evaluate NOVA on three autonomous driving benchmarks: nuScenes~\cite{caesar2020nuscenes}, V2X-Seq-SPD~\cite{yu2023v2x}, and KITTI~\cite{geiger2012we}. 
To facilitate open-vocabulary tracking, categories are split into \emph{Base}  and \emph{Novel}. 
For V2X-Seq-SPD~\cite{yu2023v2x}, Car, Van, Pedestrian, and Motorcyclist are treated as \emph{Base} classes, whereas Bus, Truck, Cyclist, and Tricyclist are \emph{Novel}. 
For KITTI~\cite{geiger2012we}, Car and Cyclist are \emph{Base} classes, with Pedestrian as \emph{Novel}. 
For nuScenes~\cite{caesar2020nuscenes}, Car, Trailer, Pedestrian, and Bicycle are \emph{Base} classes, whereas Truck, Bus, and Motorcycle are \emph{Novel}.

We adopt \emph{(s)AMOTA}~\cite{weng2020ab3dmot,caesar2020nuscenes} as the primary metric, alongside AMOTP, MOTA, MOTP, and MT~\cite{ding20233dmotformer,meinhardt2022trackformer}, reporting sAMOTA when available and AMOTA otherwise. 
A unified 3D IoU threshold of 0.25 is used for matching across all categories, and results are reported separately for Base and Novel classes.

\subsubsection*{\textbf{Implementation Details}}

Open-vocabulary 3D detections are obtained using Find n\textquotesingle{} Propagate~\cite{etchegaray2024find} on nuScenes~\cite{caesar2020nuscenes} and its upgraded variants with GroundingDINO~\cite{liu2024grounding} and YOLO-World~\cite{cheng2024yolo} on KITTI~\cite{geiger2012we} and V2X-Seq-SPD~\cite{yu2023v2x}. NOVA employs Qwen2.5-0.5B as the backbone, fine-tuned with LoRA. 
All experiments are conducted on RTX 3090~GPUs.

\subsection{Main Results and Analysis}

Table~\ref{tab:main_result} reports the main results under the Base/Novel splits on three OV-3D-MOT benchmarks. 
On \textbf{nuScenes}~\cite{caesar2020nuscenes}, NOVA substantially improves \emph{novel}-category tracking over Open3DTrack: Novel AMOTA increases from 2.20 to 22.41 and Novel MOTA from 2.70 to 19.60, indicating much stronger identity preservation when open-vocabulary semantics are uncertain. Base performance is comparable, with slightly lower AMOTA but similar MOTA/MOTP trends, suggesting that the main benefit comes from handling novel-category instability rather than over-tuning to base classes.

On \textbf{V2X-Seq-SPD}~\cite{yu2023v2x}, NOVA consistently outperforms Open3DTrack~\cite{ishaq2025open3dtrack} across two upstream OV detectors. With Find n\textquotesingle{} Propagate~\cite{etchegaray2024find} + GroundingDINO~\cite{liu2024grounding}, NOVA improves Base/Novel sAMOTA from $26.50/11.07$ to $68.17/22.95$, and raises Novel MOTA/MT from $0.59/0.23$ to $5.30/3.94$. With Find n\textquotesingle{} Propagate~\cite{etchegaray2024find} + YOLO-World~\cite{cheng2024yolo}, NOVA also improves Base sAMOTA from $29.13$ to $58.88$ and maintains competitive Novel sAMOTA. Although Novel AMOTA remains negative ($-141.78$ and $-129.85$), this mainly reflects AMOTA's sensitivity to accumulated FP, FN, and IDS caused by noisy novel-category proposals. Since NOVA improves association rather than upstream 3D proposal generation, detection-induced errors can still dominate AMOTA under low-quality open-vocabulary detections.

For \textbf{KITTI}~\cite{geiger2012we}, we only report NOVA.
In our experiments, Open3DTrack~\cite{ishaq2025open3dtrack} on KITTI~\cite{geiger2012we} produced degenerate tracking performance, likely due to the limited training size and its reliance on stable proposal behavior; hence its results are omitted.
In contrast, NOVA achieves strong Base performance and meaningful Novel tracking, highlighting improved robustness when scaling OV-3D-MOT to smaller benchmarks.

\subsection{Ablation Studies}

\subsubsection*{\textbf{Impact of LLM Model}}
Table~\ref{tab:backbone_ablation} compares LLM backbones from $0.5$B to $3.8$B across the Qwen, Llama, and Phi families. We observe that simply scaling model size does not consistently improve tracking performance. \textbf{Qwen2.5-0.5B} achieves the highest sAMOTA of $45.56\%$ and the best Novel sAMOTA of $22.95\%$, suggesting that a compact model is sufficient to capture geometric and motion priors while avoiding excessive semantic bias. Larger models, such as Llama-3.2-3B and Qwen2.5-1.5B, reduce IDS and FP but produce more FN, indicating a more conservative association strategy. In contrast, Qwen2.5-0.5B maintains higher recall, recovers more difficult tracks, and runs at \textbf{$3.4$ FPS}, nearly $3\times$ faster than Llama-3.2-3B and $4\times$ faster than Phi-3.5-mini. Overall, Qwen2.5-0.5B provides the best balance between tracking accuracy and efficiency.

\begin{table}[t]
    \centering

    \caption{Ablation on Hybrid Prompting Strategy.}
    \label{tab:ablation_open_vocab}
    \vskip-1ex
    \small 
    \setlength{\tabcolsep}{3pt}
    
    \begin{tabularx}{0.98\columnwidth}{l c Y Y Y}
        \toprule
        \multirow{2}{*}{\textbf{Strategy}} & \textbf{Training Handling} & \multicolumn{3}{c}{\textbf{sAMOTA} $\uparrow$} \\
        \cmidrule(lr){3-5} 
        & \textbf{of Novel Classes} & \textbf{All} & \textbf{Base} & \textbf{Novel} \\
        \midrule
        Baseline & Ignored / Discarded & 38.55 & 62.46 & 14.63 \\
        
        \rowcolor{headergray} \textbf{Ours} & \textbf{Labeled as ``Unknown''} & \textbf{45.56} & \textbf{68.17} & \textbf{22.95} \\
        
        \bottomrule
    \end{tabularx}
    \vskip-3ex
\end{table}

\subsubsection*{\textbf{Analysis of the Geometry Encoding Strategy}}
Table~\ref{tab:ablation_geometry} reveals two takeaways.
First, representing 3D states as learned geometry tokens is \emph{necessary} for open-vocabulary association: compared with text-only coordinates, the Geometry Encoder primarily improves the \emph{novel} split while keeping the base split essentially saturated, indicating that novel tracking is limited by noisy geometry/semantics where tokenization-sensitive numeric strings fail to provide stable cues.
Second, the IoU-quality auxiliary supervision acts as a \emph{quality-aware regularizer}: it yields only a small sAMOTA gain over geometry encoding alone, but improves error-driven behavior (higher MOTA and fewer FNs), suggesting the model becomes more conservative against low-fidelity detections and forms longer, cleaner trajectories rather than merely trading errors.
Overall, the table supports our design choice that geometry-aware injection is the key enabler, while quality supervision further stabilizes association under open-vocabulary noise.

\subsubsection*{\textbf{Ablation on Open-Vocabulary Training Strategy}}
Table~\ref{tab:ablation_open_vocab} demonstrates the efficacy of incorporating novel objects during training. The baseline approach, which discards novel instances, suffers from overfitting to specific semantic priors, resulting in poor generalization to unseen categories (14.63\% Novel sAMOTA). In contrast, labeling these instances as a generic \texttt{Unknown} token encourages the model to learn \textit{class-agnostic association cues} rather than relying on class-specific visual patterns. This strategy significantly boosts Novel sAMOTA by \textbf{+8.32\%} (to 22.95\%). 
Furthermore, it unexpectedly improves Base category performance by \textbf{+5.71\%} (to 68.17\%), as the increased geometric diversity of \texttt{Unknown} samples serves as a robust regularizer against~overfitting.

\begin{table}[!t]
    \centering
    \caption{Ablation on Data Sampling Strategies.}
    \label{tab:ablation_sampling_full}
    \vskip-1ex
    \small 
    \setlength{\tabcolsep}{2pt}
    
    \begin{tabularx}{0.98\columnwidth}{l c c YYYY}
        \toprule
        \multirow{2}{*}{\textbf{Method}} & \textbf{Neg.} & \textbf{Pos.} & \textbf{sAMOTA} & \textbf{AMOTA} & \textbf{MOTA} & \textbf{MT} \\
        & \textbf{Samp.} & \textbf{Jitter} & $\uparrow$ & $\uparrow$ & $\uparrow$ & $\uparrow$ \\
        \midrule
        
        Baseline & Random & $\checkmark$ & 41.46 & -67.77 & 18.26 & 6.86 \\
        
        Baseline$^\dagger$ & Local & $\checkmark$ & 42.72 & \textbf{-57.40} & 19.48 & 8.17 \\
        
        \midrule
        
        Variant & Hard & $\times$ & 43.23 & -70.26 & 19.38 & 8.43 \\
        
        \rowcolor{headergray} \textbf{Ours} & \textbf{Hard} & \textbf{\checkmark} & \textbf{45.56} & -64.36 & \textbf{20.30} & \textbf{9.16} \\
        
        \bottomrule
    \end{tabularx}
    \vskip-2ex
\end{table}
\begin{table}[!t]
    \centering
    \caption{Ablation on Inference History Length.}
    \label{tab:ablation_history_len}
    \vskip-1ex
    \small 
    \setlength{\tabcolsep}{5pt}
    
    \begin{tabularx}{0.98\columnwidth}{c YY cc}
        \toprule
        \textbf{History} & \multicolumn{2}{c}{\textbf{sAMOTA} $\uparrow$} & \textbf{MOTA} $\uparrow$ & \textbf{IDS}$\downarrow$ \\
        
        \cmidrule(lr){2-3} 
        
        \textbf{Length ($L$)} & \textbf{All} & \textbf{Novel} & \textbf{All} & \textbf{All} \\
        \midrule
        
        0 & 0.00 & 0.00 & -12.05 & 1785 \\
        1 & 43.17 & 16.81 & 16.71 & 155 \\
        
        \rowcolor{headergray} \textbf{3 (Default)} & \textbf{45.56} & \textbf{22.95} & 20.30 & 182 \\
        
        5 & 41.25 & 14.56 & \textbf{20.86} & \textbf{128} \\
        
        \bottomrule
    \end{tabularx}
    \vskip-5ex
\end{table}
\begin{figure*}[!t]
    \vspace*{1mm}
    \centering
    \includegraphics[width=1.0\textwidth]{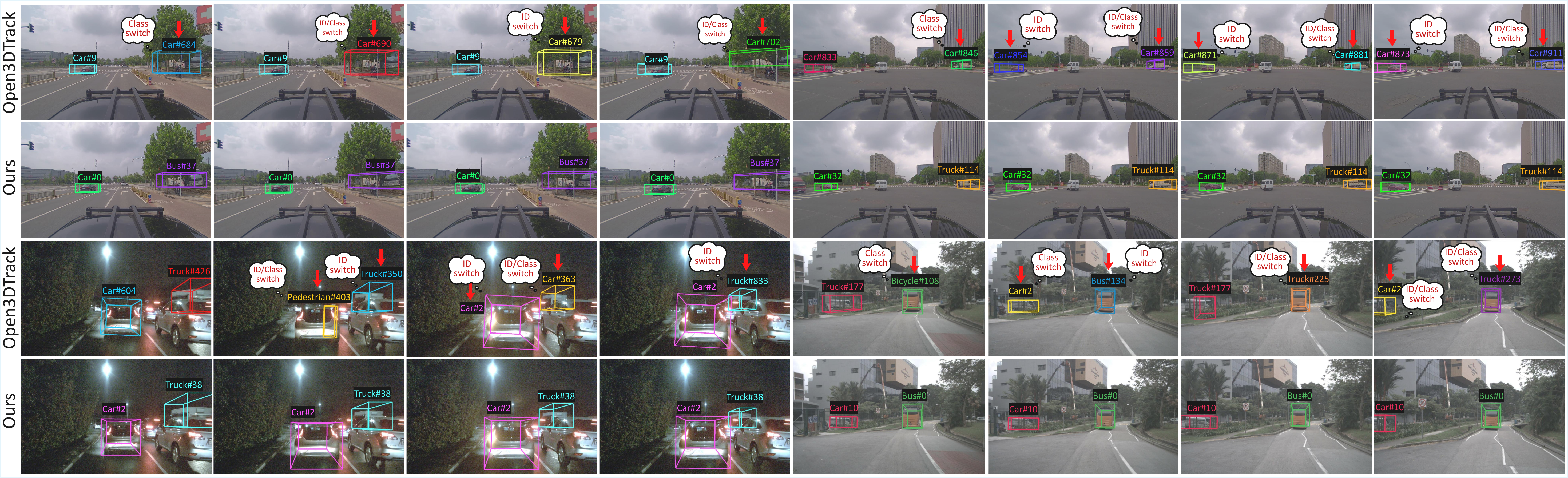} 
    \vskip-1ex
    \caption{\textbf{Qualitative comparison of OV-3D-MOT across datasets.}
    Four representative scenes from different autonomous-driving benchmarks, each visualized over four consecutive frames ($t{=}1{\sim}4$).
    For each scene, we compare Open3DTrack~\cite{ishaq2025open3dtrack} (top) with NOVA (bottom).
    Callouts and arrows highlight typical open-vocabulary tracking failure cases, including \emph{Class switch}, \emph{ID switch}, and \emph{ID/Class switch}.}
    \label{fig:vis}
    \vskip-3ex
\end{figure*}

\subsubsection*{\textbf{Ablation on Data Sampling Strategies}}
Table~\ref{tab:ablation_sampling_full} analyzes the construction of positive and negative training pairs. Negative-sample quality is crucial: Random sampling gives the lowest sAMOTA of $41.46\%$, since most negatives are spatially distant and easy to distinguish. Restricting negatives to a Local neighborhood improves sAMOTA to $42.72\%$, but remains inferior to Hard Negative Mining. By focusing on confusing nearby objects (\textit{e.g.}, adjacent boxes with high overlap), our strategy learns finer discriminative cues and raises sAMOTA to \textbf{$45.56\%$}.
Positive-sample augmentation is also important. Training without jitter overfits to ideal ground-truth centers and performs worse, with $43.23\%$ sAMOTA. Adding geometric jitter simulates detector localization noise, improving robustness to imperfect proposals and increasing sAMOTA by $+2.33\%$ while raising MT to $9.16\%$.

\subsubsection*{\textbf{Ablation on Inference History Length}}
We analyze the impact of inference history length ($L$) in Table~\ref{tab:ablation_history_len}.
Without history ($L=0$), performance collapses to 0.00\% sAMOTA, confirming the necessity of temporal context.
Introducing just one frame ($L=1$) boosts sAMOTA to 43.17\%, establishing basic temporal consistency.
Extending history to three frames achieves the best trade-off, peaking on All (45.56\%) and Novel (22.95\%) by capturing short-term motion dynamics.
Further increasing $L$ to 5 leads to a performance drop (41.25\%): while longer history reduces identity switches, it also induces an overly conservative bias that hurts recall on novel categories.
We therefore set $L=3$ by default.

\begin{table}[t]
    \centering
    \caption{Impact of Training Epochs.}
    \label{tab:ablation_epochs}
    \vskip-1ex
    \small 
    \setlength{\tabcolsep}{4pt}

    \begin{tabularx}{0.98\columnwidth}{c YY cc} 
        \toprule
        \multirow{2}{*}{\textbf{Epochs}} & \multicolumn{2}{c}{\textbf{sAMOTA} $\uparrow$} & \textbf{MOTA} $\uparrow$ & \textbf{MT}$\uparrow$ \\

        \cmidrule(lr){2-3} 
        
         & \textbf{All} & \textbf{Novel} & \textbf{All} & \textbf{All} \\
        \midrule
        1  & 37.09 & 12.11 & 17.64 & 7.95 \\
        5  & 43.79 & 18.52 & 21.27 & \textbf{9.18} \\
        
        \rowcolor{headergray} \textbf{8 (Default)} & \textbf{45.56} & \textbf{22.95} & 20.30 & 9.16 \\
        
        10 & 44.49 & 20.74 & \textbf{21.30} & 8.41 \\
        12 & 43.91 & 18.62 & 20.45 & 8.49 \\
        15 & 44.20 & 18.80 & 20.14 & 9.03 \\
        \bottomrule
    \end{tabularx}
    \vskip-3ex
\end{table}

\subsubsection*{\textbf{Impact of Training Epochs}}
As shown in Table~\ref{tab:ablation_epochs}, performance peaks at the 8th epoch, achieving the highest sAMOTA on All (45.56\%) and Novel (22.95\%).
Extending training to 10 epochs slightly increases MOTA (21.30\% \textit{vs.} 20.30\%), but Novel sAMOTA drops to 20.74\%, indicating reduced open-vocabulary generalization.
Further training (12--15 epochs) amplifies this trend, suggesting overfitting to Base categories.
We therefore adopt 8 epochs to balance accuracy and open-vocabulary generalization.

\subsection{Qualitative Analysis}
As shown in Fig.~\ref{fig:vis}, Open3DTrack~\cite{ishaq2025open3dtrack} exhibits typical failure modes under open-vocabulary uncertainty, including \emph{class switch}, \emph{ID switch}, and their combinations. Such errors are more likely when novel-category detections are semantically ambiguous and geometrically jittery, making cost-based association overly sensitive to short-term fluctuations. Failures become pronounced in crowded scenes with adjacent objects and partial overlap, under challenging visibility where box stability and confidence are inconsistent, and for distant objects where sparse measurements further degrade localization and semantic reliability.

NOVA is more robust by making association a trajectory-conditioned decision rather than a frame-wise similarity snapshot. Geometry-aware Embeddings retain fine-grained 3D cues, Hybrid Prompting reduces dependence on unreliable novel-category names, and Hard Negative Mining strengthens discrimination among close competitors. Together, these designs reduce sensitivity to transient semantic noise and localization jitter, producing more stable identities across frames. Such robustness is important for autonomous driving, where planning and forecasting rely on temporally consistent tracks, particularly in dense traffic or adverse conditions.

\section{Conclusion}
We presented NOVA, an autoregressive association formulation for open-vocabulary 3D multi-object tracking within the tracking-by-detection framework. By leveraging an LLM as a trajectory-context model, NOVA converts association into next-token decision making, enabling geometry-aware embeddings and controlled semantic cues to be jointly modeled under open-vocabulary uncertainty. This also suggests that large models hold considerable potential for open-vocabulary perception, where flexible semantic generalization and context-aware reasoning are essential for handling unknown categories. Beyond quantitative results, our qualitative analysis shows that such context-conditioned association helps reduce class and ID switches in crowded scenes and under partial occlusion, where short-term geometry or category cues are often unreliable. Nevertheless, severe occlusion and highly dense interactions remain challenging when upstream proposals are fragmented or missing. Future work will explore lightweight appearance cues, stronger temporal memory, and more efficient large-model adaptation, further investigating generative modeling for robust open-world autonomous driving perception.

\section*{Reference} 
\begingroup
    \renewcommand{\section}[2]{} 
    \bibliographystyle{IEEEtran}
    \bibliography{references} 
\endgroup
\end{document}